%% file: main2.tex
\documentclass[10pt,twocolumn,letterpaper]{article}

\usepackage{cvpr} 

\usepackage{graphicx}
\usepackage{amsmath}
\usepackage{amssymb}
\usepackage{booktabs}
\usepackage{xcolor, colortbl}
\usepackage[accsupp]{axessibility}
\usepackage{times}
\usepackage{latexsym}
\usepackage{float}
\usepackage{makecell}
\usepackage{comment}
\usepackage{booktabs}
\usepackage{multirow}

\usepackage{amsfonts}
\usepackage{bbm}
\usepackage{multicol}

\definecolor{LightGrey}{rgb}{0.9,0.9,0.9}

\usepackage[pagebackref,breaklinks,colorlinks]{hyperref}

\usepackage[capitalize]{cleveref}
\crefname{section}{Sec.}{Secs.}
\Crefname{section}{Section}{Sections}
\Crefname{table}{Table}{Tables}
\crefname{table}{Tab.}{Tabs.}

\usepackage{times}
\usepackage{epsfig}

\usepackage[font=small,labelfont=bf,tableposition=top]{caption}

\usepackage{xcolor}

\usepackage{enumitem}
\setlist[itemize]{%
topsep=5pt,
labelsep=5pt,%
labelindent=0.4\parindent,%
itemindent=0pt,%
leftmargin=*,%
itemsep=-1pt
}

\newcommand{\transfo}[2]{\textit{#1}$\rightarrow$\textit{#2}}
\newcommand{\wi}{w_1 \rightarrow w_2}
\newcommand{\p}{\mathbb{P}_{O}}

\usepackage[T1]{fontenc}

\usepackage[utf8]{inputenc}

\usepackage{microtype}

\begin{document}

\title{Embedding Arithmetic of Multimodal Queries for Image Retrieval}

\def\cvprPaperID{****} 
\def\confName{CVPR}
\def\confYear{2022}

\author{Guillaume Couairon\\
Meta AI, Sorbonne Université\\

{\tt\small gcouairon@fb.com}
\and
Matthijs Douze \\
Meta AI\\
{\tt\small matthijs@fb.com}
\and
Matthieu Cord \\
Valeo, Sorbonne Université \\
{\tt\small matthieu.cord@sorbonne-universite.fr}
\and
Holger Schwenk \\
Meta AI\\
{\tt\small schwenk@fb.com}
}

\maketitle
\input{abstract}

\input{intro.tex}

\input{related.tex}

\input{dataset.tex}

\input{methods.tex}

\input{experiments.tex}

\input{conclusion.tex}

{\small
\bibliographystyle{ieee_fullname}
\bibliography{anthology,custom}
}


\end{document}

%% file: abstract.tex
\begin{abstract}

Latent text representations exhibit geometric regularities, such as the famous analogy: \textit{queen} is to \textit{king} what \textit{woman} is to \textit{man}.
Such structured semantic relations were not demonstrated on image representations.
Recent works aiming at bridging this semantic gap embed images and text into a multimodal space, enabling the transfer of text-defined transformations to the image modality.

We introduce the SIMAT dataset to evaluate the task of Image Retrieval with Multimodal queries. SIMAT contains 6k images and 18k textual transformation queries that aim at either replacing scene elements or changing pairwise relationships between scene elements. The goal is to retrieve an image consistent with the (source image, text transformation) query.
We use an image/text matching oracle (OSCAR) to assess whether the image transformation is successful.
The SIMAT dataset will be publicly available.

We use SIMAT to evaluate the geometric properties of multimodal embedding spaces trained with an image/text matching objective, like CLIP. We show that vanilla CLIP embeddings are not very well suited to transform images with delta vectors, but that a simple finetuning on the COCO dataset can bring dramatic improvements.
We also study whether it is beneficial to leverage pretrained universal sentence encoders (FastText, LASER and LaBSE).

\end{abstract}

%% file: intro.tex
\begin{figure}[t!]
    \centering
    \includegraphics[width=\linewidth]{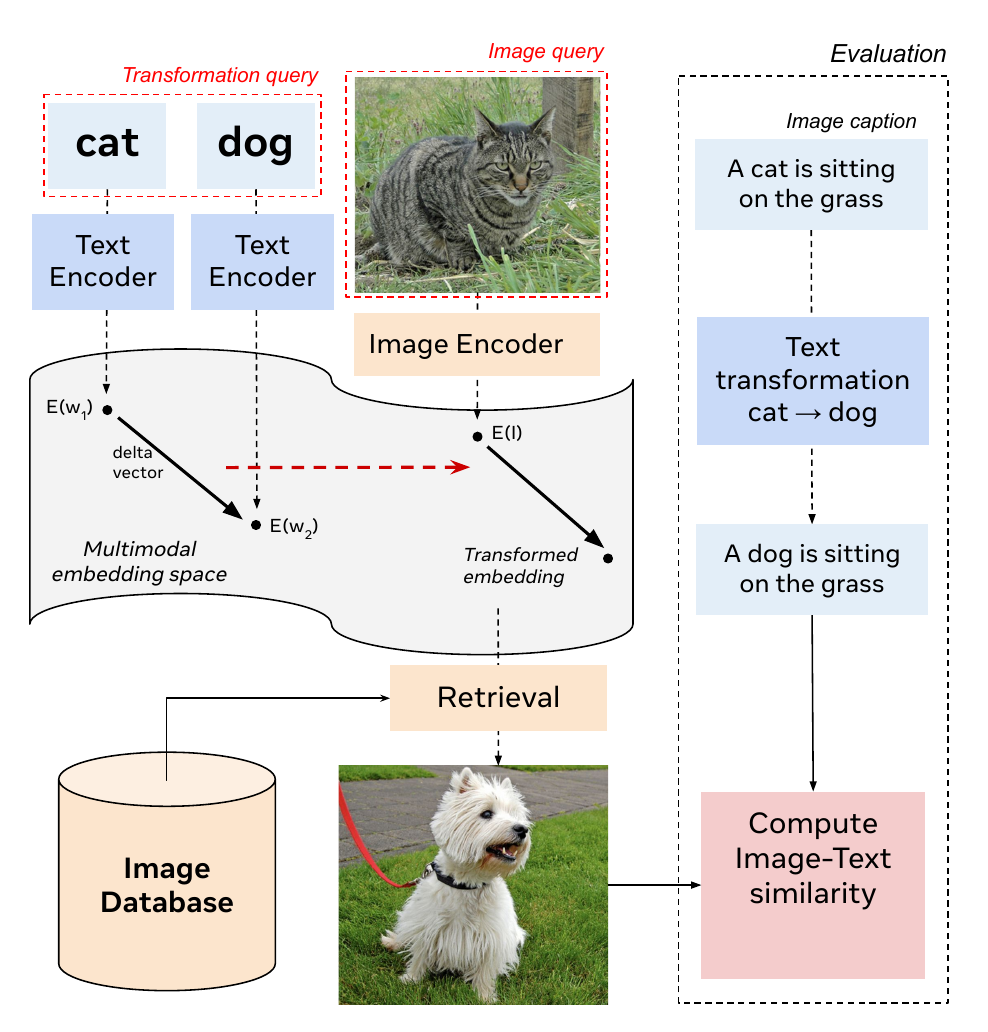}
    \caption{
    We aim at performing image retrieval, guided by an input image and a text query describing how the input image should be transformed. The transformation is mapped to a \textit{delta vector}, added to the image embedding to produce a \textit{transformed embedding}, for which a corresponding image is retrieved in a database. The evaluation module checks that the text-transformed caption is valid for the image result.
    }
    \label{fig:splash}
    \vspace{-1em}
\end{figure}

\section{Introduction}

%
Many works aim at learning a multimodal space in which images and text can be embedded and compared \cite{ wang2016learning, faghri2018vse++, engilberge2018finding, zheng2020dual}. This is especially useful for image/text retrieval, where given a sample from one modality, one has to find a corresponding sample of the other modality in a database.
Such embeddings have other interesting applications. It has been shown in \cite{vo2019let, jia2021scaling} that textual input can be used to \emph{replace} scene elements in images or change their properties with simple arithmetic operations in the latent space, combined with nearest neighbour search.

In this work, we study the following problem: given an image and a transformation query formulated in the text domain, the task is to find an image that satisfies the transformation query while being semantically as similar as possible to the input image.

For example, with the \transfo{cat}{dog} transformation, an image showing a cat sitting in the grass should be transformed into an image with a dog sitting in the grass (Figure~\ref{fig:splash}). 

\textbf{Evaluation challenges.} We seek to create a dataset to evaluate algorithms on this task. Such a dataset should contain feasible (image, text transformation) queries: the transformation \transfo{man}{dog} can be applied to an image with \textit{``A man is running on the beach''}, but not to \textit{``A man is speaking on the phone''}. We create a corpus from Visual Genome images and annotations \cite{krishna2017visual} and ensure that this requirement is met.
Evaluating semantic text-to-image transfer is challenging: First, we need to ensure that the requested transfer was performed (the cat was replaced by a dog).
Then, we need to verify that the modification is minimal: the dog should be sitting on grass and ideally, all other visual elements should not be changed. We use OSCAR \cite{li2020oscar} as an external oracle to assess whether these two conditions are met. OSCAR has been trained on captioned images with a binary cross-entropy loss to recognize whether or not a given text corresponds to an image.



\textbf{Embedding arithmetic of multimodal queries.} We choose to transform images by encoding the transformation query as a \textit{delta vector} in the multimodal space, before adding it to an image embedding and retrieving the closest image in a database (see Figure \ref{fig:splash}). This operation solely relies on the image/text alignment without needing any transformation example. However, it requires a well structured multimodal space to be able to transfer text transformations to images. We know that word and sentence embeddings trained on vast amounts of data have been shown to possess geometric properties that can be useful for text transformation \cite{mikolov2013efficient,logeswaran2018efficient}. Previous work \cite{jia2021scaling} has hinted that such geometric properties could also be present in multimodal spaces, without quantitative evidence. 

In this paper, we study the suitability of multimodal embedding spaces like CLIP \cite{radford2021learning} to perform image retrieval with image-text queries. The CLIP multimodal space was trained with a contrastive image/text matching loss \cite{infonce} on 400M image-text pairs crawled from the internet, which makes it suitable for use in a zero-shot evaluation setting like ours. Given that word and sentence embeddings have been shown to display strong geometric properties \cite{mikolov2013efficient,logeswaran2018efficient}, we study whether it is necessary to use them as building blocks to multimodal embeddings better suited to image retrieval with multimodal queries. In particular we use LASER \cite{artetxe2019massively} and LaBSE \cite{feng2020language}, which have been pre-trained on large corpora of multilingual data.

The contributions of this work are:
\begin{itemize}
    \item SIMAT, a dataset of 6~000 images and 18~000 transformation queries, to evaluate algorithms on the task of image retrieval with multimodal queries, which comes with an evaluation metric based on OSCAR.
    
    \item detailed experiments to measure which multimodal embeddings work best within the \textit{delta vectors} framework to transform images.
\end{itemize}


%% file: related.tex
\pagebreak
\section{Related Work}

The geometric properties of word embeddings have been observed notably in \cite{mikolov2013efficient, fournier2020analogies}, with the famous example \textit{king} is to \textit{queen} what \textit{man} is to \textit{woman}. These properties have also been studied for sentence embeddings \cite{logeswaran2018efficient}.
Recent work has demonstrated that state-of-the art multimodal embeddings can be obtained by scaling up image-text alignment pretraining \cite{jia2021scaling,radford2021learning}. It has been observed that such embeddings exhibit properties similar to word embedding analogies, allowing textual and image queries to be combined. A few examples are shown in \cite{jia2021scaling}, but without any quantified evaluation.


\textbf{Image Retrieval with Multimodal Queries.} In general, Image Retrieval with Multimodal Queries is a form of image retrieval where some textual inputs serve as instructions to modify an existing image through retrieval. The instruction is a simple word pair in our case, but it can also be a more complex instruction in natural language.
%
Several datasets exist to tackle this problem: The CSS dataset \cite{vo2019composing} which is a synthetic dataset with simple colored geometrical objects based on CLEVR \cite{johnson2017clevr}. The Fashion200k dataset \cite{han2017automatic} provides around 200k images of fashion products, each annotated with a compact attribute description. Similarly, the Fashion-IQ dataset \cite{guo2019fashion} was built to advance research on interactive fashion image retrieval. The MIT-States dataset \cite{isola2015discovering}, also commonly used, is a dataset of $\approx$60k images, each annotated with an object/noun label and a state/adjective label such as \textit{new car} or \textit{broken window}.
%
Those datasets are designed to evaluate image retrieval with multimodal queries on narrow domains, which gives more control over what attributes can be changed and ensures that the transformation is always feasible. Another common characteristic is the focus on changing object properties rather than objects relationships. We focus on more realistic images, and study object transformations where an object should be replaced by another without changing the high-level subject-object interaction.

%
Methods for solving this image retrieval task \cite{vo2019composing, anwaar2021compositional, song2019polysemous} all focus on supervised learning: a fraction of the dataset is used for training and the remaining for testing.
Instead, we want to measure if multimodal embeddings trained with an image/text matching objective can be used to solve this task without any transformation example.

\textbf{Text-driven Image Editing.} Some work focus on directly modifying the pixels of query images instead of performing the retrieval step \cite{shi2020benchmark}. \cite{zheng2020semantic} encode images as a graph of interacting objects which lets the user modify an image by editing its scene graph. GANs are also frequently used to modify images based on some natural language input \cite{nam2018text, li2020manigan, xia2020tedigan}. Lastly, CLIP \cite{radford2021learning} can be used in combination with a StyleGAN generator to make semantic edits in images, as exemplified in \cite{patashnik2021styleclip}.



%% file: dataset.tex
\section{The SIMAT database}
\label{sec:simat}

\subsection{Requirements}
\label{sec:data:eval}

First, we need a list of images with some transformation queries (e.g. a man sleeping on the beach, with the query \transfo{man}{woman}). We want simple images (so that the query is unambiguous) and relevant transformation queries.
Second, we need a database of images that we will use for the retrieval step.
And finally, we need a criterion to decide, based on the retrieved image, if the transformation is successful or not. It is the case if only the element designated by the transformation query has changed, while keeping the rest of visual elements as similar as possible.

Previous work \cite{vo2019let} has tried to solve these requirements with a dataset of $\approx$1,500 images from Google Image Search queries, dubbed SIC112, with each image annotated with a actor-action-environment triplet, such as \textit{(woman, walking, street)}, among a set of 112 possible triplets. Transformation queries then consist in changing either the subject, action or environment. 
This set of images is also used as a database for retrieval, which has two advantages: (i) transformation queries are always possible by design of the dataset, and (ii) the quality of the retrieved image is measured by checking if its annotation triplet was indeed the one expected by the transformation query.

We scale this approach to a larger number of annotation triplets, that take the more general form of (subject, relationship, object).
However, we observed that due to the larger triplet vocabulary, images can be accurately described by multiple such triplets, which skews the evaluation metric: an image would often be rejected for not being annotated with the expected triplet while still being visually correct. 
Therefore, we chose to use a different metric for evaluating the quality of transformed images: we evaluate whether the semantic transformation is successful by querying OSCAR \cite{li2020oscar}. OSCAR computes the probability $\p(I, T)$ that a caption $T$ accurately described an image $I$, based on the concatenation of the text tokens in $T$ and the object tags and features detected by faster R-CNN on image $I$ (we provide the triplet to OSCAR in the form of a caption written in natural language). Note that this OSCAR-based evaluation method does not involve image annotations in the retrieval database and thus could potentially be applied to a much larger database of non-annotated images, which we leave for future work.

\subsection{Construction}
\label{sec:simat:constr}

\begin{figure}
    \centering
    \includegraphics[width=\linewidth]{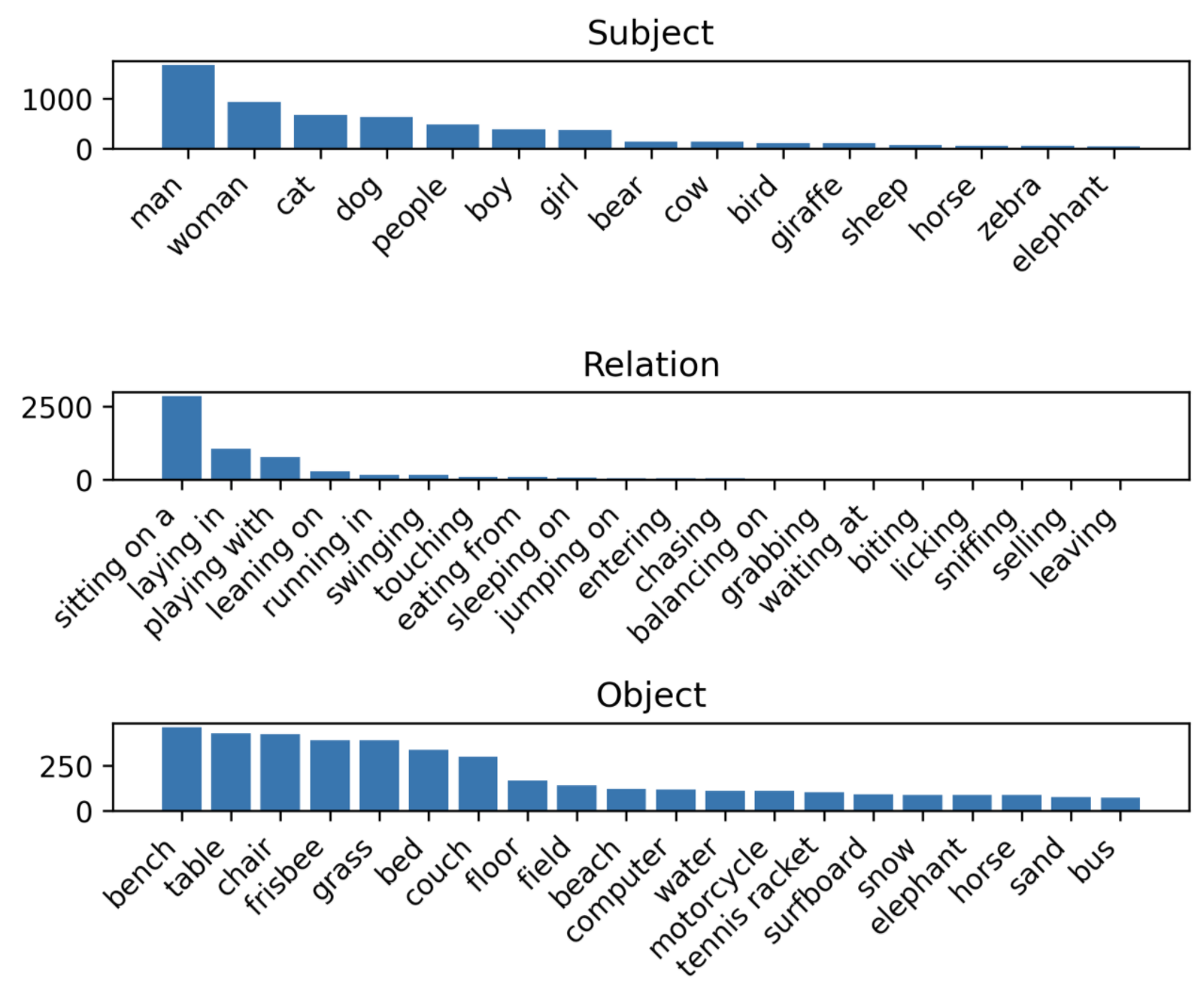}
    \caption{Statistics for SIMAT database. All subjects and relationships are represented, but only 25 objects out of 131 are listed here.}
    \label{fig:simat_stats}
\end{figure}

Similarly to \cite{vo2019let}, we create a list of images annotated with (subject, relationship, object) triplets, and
perform the retrieval step inside the same list of images to ensure that transformation queries always have a valid solution in the dataset.
We start from annotations from the Visual Genome dataset \cite{krishna2017visual}. Each image in the dataset contains a list of such triplets with subject and object bounding boxes, which we use to crop square regions of images that minimally contain the subject and object in the image.
We then filter this list and compute possible transformations:

\textbf{Subject/Relation filtering.} Only keep triplets for which the subject is a human or animal, and the relation is a non-positional relationship in Visual Genome. The full lists are shown in Figure \ref{fig:simat_stats}.

\textbf{Object filtering.} Only keep objects $O$ for which there exists at least two triplets $(S, R, O)$ and $(S', R', O)$ with $R \neq R'$. This ensures that the selected objects have at least two different types of interaction in images. Then, only keep the 10 most frequent objects for a single (subject, relation) pair. This gives a list of 645 distinct triplets.

\textbf{Building transformation queries.} For each image $I$ with associated triplet $(S, R, O)$, add in the list of transformation queries $(I, O\rightarrow O')$ if there is a triplet $(S, R, O')$ in the database. Do the same for $S$ and $R$. This ensures that transformation queries consists of pairs of \textit{objects} that can have the same (subject, relation) pair, and symmetrically for \textit{subjects} and \textit{relations}.

\textbf{Writing captions for OSCAR.} For each of the 645 triplets, we manually wrote a caption in natural language, e.g. (\textit{man}, \textit{sitting on}, \textit{chair}) $\rightarrow$ \textit{A man sitting on a chair}.

We now have a database of images and transformation queries, but we have noticed some noise in the annotation procedure: an image can have a triplet annotation which does not well describe the main action in it, because the cropping procedure included an object that is more important than the extracted triplet. Also, transformation queries sometimes consisted in synonyms.
We solve this problem using OSCAR to filter transformation queries: Given an image $I$ with query triplet $t_1$ and target triplet $t_2$, we keep the corresponding transformation if $\p(I, t_1) > 0.9$ and $\p(I, t_2) < 0.1$. This ensures that not modifying the image is not a valid solution to the problem.

The distribution of images being quite skewed (see Figure~\ref{fig:simat_stats}), the transformation queries also have a bias towards the more frequent subjects, relations and objects. We alleviate this problem by using reweighting in the scoring metric (see below).

In summary, our SIMAT dataset (for Semantic IMage Transformation) consists of:
\begin{itemize}
    \item 5~989 images, each annotated with a subject-relation-object triplet. 
    \item 17~996 transformation queries on those images, with queries asking to change the subject, the relation, or the object.
    \item A list of 645 distinct subject-relation-object triplets with corresponding captions, each triplet having at least 2 corresponding images. 
\end{itemize}

To allow hyperparameter selection, 
we make a 50-50 dev/set split on the list of images, and split the transformation queries accordingly.

\subsection{Evaluation Metric}
\label{sec:simat:eval}

Let $(I_i, \wi, T_i)$ be a sample in our dataset where $\wi$ is the transformation query and $T_i$ is the caption associated to the target triplet of this sample. For this sample, we consider that a retrieved image $J_i$ corresponds to a successful transformation if OSCAR outputs a probability $\p(J_i, T_i) > 0.5$.
The final score is simply a weighted accuracy over all dataset samples:
\begin{equation}
S = \sum_{(I_i, T_i, \mu_i) \in S} \mu_i \mathbbm{1}_{\p(J_i, T_i)>0.5}
\end{equation}
where the coefficients $\mu_i$ are the contributions of each sample to the total score. We adopt an inverse square root reweighting to downsample the most frequent transformations.

%% file: methods.tex
\section{Methods}
\label{sec:methods}

Starting from semantic transformations in text, we show how text transformations can be transferred to images via multimodal embeddings. We then present our procedure for fine-tuning multimodal embeddings.

\subsection{Text delta vectors for semantic transformations}

Semantic properties in sentences can be modified by word replacement: in the sentence \textit{``A man walking on the beach''}, the semantic property \textit{subject gender} can be changed by replacing the word \textit{man} with the word \textit{woman}. 
In a latent space, where direct word replacement is not possible, we can apply semantic transformations by doing arithmetic operations. By encoding sentences as the sum of their word embeddings, applying a transformation $\wi$ on a sentence embedding $E(s)$ amounts to adding the vector $E(w_2) - E(w_1)$, which we call a \textit{delta vector}. In principle, the textual form of the transformed sentence can be found by retrieving the sentence embedding closest to $E(s) + E(w_2) - E(w_1)$ in a database. 

However, there is some ambiguity in the process since bag of words representations do not take into account the order of words. That is why we consider more complex non-linear sentence embeddings which have been shown to display similar properties as above \cite{logeswaran2018efficient}, in addition to better reflecting the meaning of sentences \cite{artetxe2019massively}.

We study four sentence embeddings: \textit{CLIP}, obtained by a contrastive loss on a large set of image/text pairs; \textit{FastText}, obtained with a weighted sum of FastText word embeddings \cite{bojanowski2017enriching}; \textit{LaBSE}, which are trained by matching parallel sentences in different languages with a contrastive loss \cite{feng2020language}; and  the \textit{LASER} embeddings \cite{artetxe2019massively} which are trained with a multilingual translation task.

\subsection{From text delta vectors to images}

Semantic transformations, seen as \textit{delta vectors} as defined above, can be added to image embeddings in multimodal spaces.
We transform images with text delta vectors in the following manner (Fig~\ref{fig:splash}): given an image encoder $E_{img}$ and a text encoder $E_{txt}$ that embed both modalities into a shared latent space, we retrieve in the SIMAT database the image that has the highest cosine similarity with the latent vector
\begin{equation}
x = E_{img}(I) + \lambda \cdot (E_{txt}(w_2) - E_{txt}(w_1))
\end{equation}
The \textit{scaling factor} $\lambda$ is a hyper-parameter that can be adjusted to increase the strength of the transformation. The natural choice is $\lambda=1$ but it has been noted that a higher value can help to better enforce the transformation \cite{jia2021scaling}.
The image embeddings are quite sparse due to the relatively small size of the image database, so we found it helpful to enforce the rule that the retrieved image should be different from the input image.

\subsection{Finetuning multimodal embeddings}

\begin{figure}[H]
    \centering
    \includegraphics[width=\linewidth]{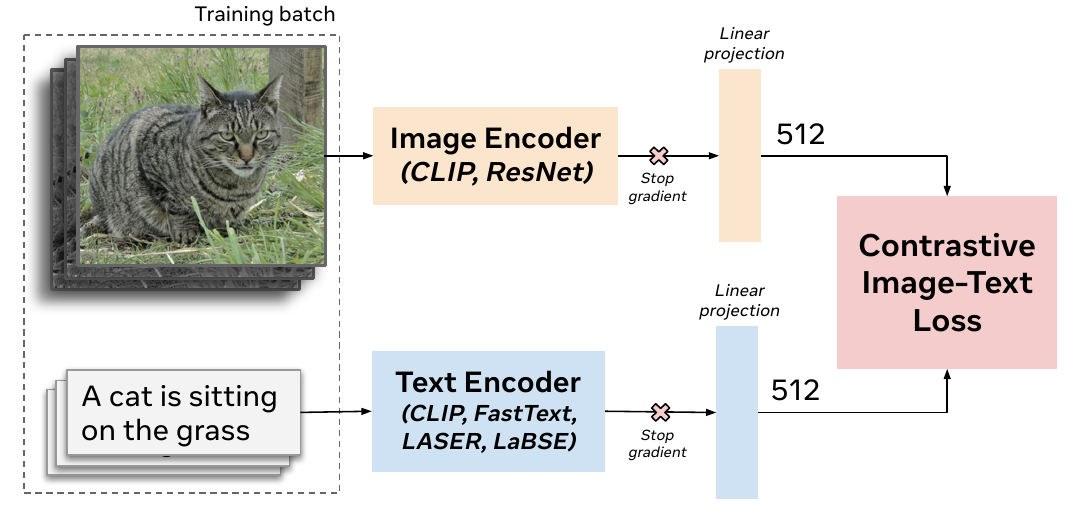}
    \caption{Layer adaptation learning on COCO. The image and text embeddings are projected to a shared multimodal space of dimension 512.}
    \label{fig:mscoco_fig}
\end{figure}

We consider multiple choices for the image and text encoders: Our default setup is to use the CLIP embeddings for both modalities (63M parameters for the text encoder, 87M for the image encoder), and we experiment with using two ImageNet-pretrained ResNets (RestNet50 and ResNet152, respectively 23M and 63M parameters) as image encoders, and FastText, LASER and LaBSE as text encoders.
We can evaluate the vanilla CLIP embeddings without retraining; however, other encoding choices are not directly compatible
and we have to fine-tune the encoders to be able to encode image and text into a shared latent space.
We use a very simple fine-tuning scheme on COCO \cite{mscoco} where we train linear adaptation heads after the frozen encoders (Fig~\ref{fig:mscoco_fig}) for 30 epochs with a learning rate of 1e-3 and a batch size of 4096. Fine-tuning a model takes approx. 3 hours on 8 Tesla V100~GPUs. 

When using the ResNet-based encoders, our initial study showed that only training a linear layer is not sufficient to get a reasonable performance on image-text retrieval, because the backbone network is only trained on image classification. Therefore, we freeze only the first three blocks of the ResNet models and add a simple 4-layer MLP architecture on top of the pooled features.
We use an image-text InfoNCE \cite{infonce} contrastive loss (which was used for training CLIP):
\begin{eqnarray}
\mathcal{C}(I, T) & = & \frac{1}{n}\sum_{i=1}^n  \Big( \frac{\exp(I_i \cdot T_i/\tau)}{\sum_{j= 1}^n {\exp(I_i \cdot T_j/\tau)}} \Big)
\nonumber
\\
\mathcal{L} & = & \frac{1}{2} \mathcal{C}(I, T) + \frac{1}{2} \mathcal{C}(T, I)
\label{eq:loss}
\end{eqnarray}
where $I$ and $T$ are normalized image and text embeddings, $\tau$ a temperature parameter which is learnable in CLIP. However, we choose to keep it fixed to study its impact on the transformation score.

%% file: experiments.tex
\section{Experiments on SIMAT database}
\label{sec:exps}

In this section, we analyze the ability of various multimodal embeddings to transfer text transformations to images via \textit{delta vectors}. 

\begin{figure*}[!ht]
    \centering
    \includegraphics[width=1.0\textwidth]{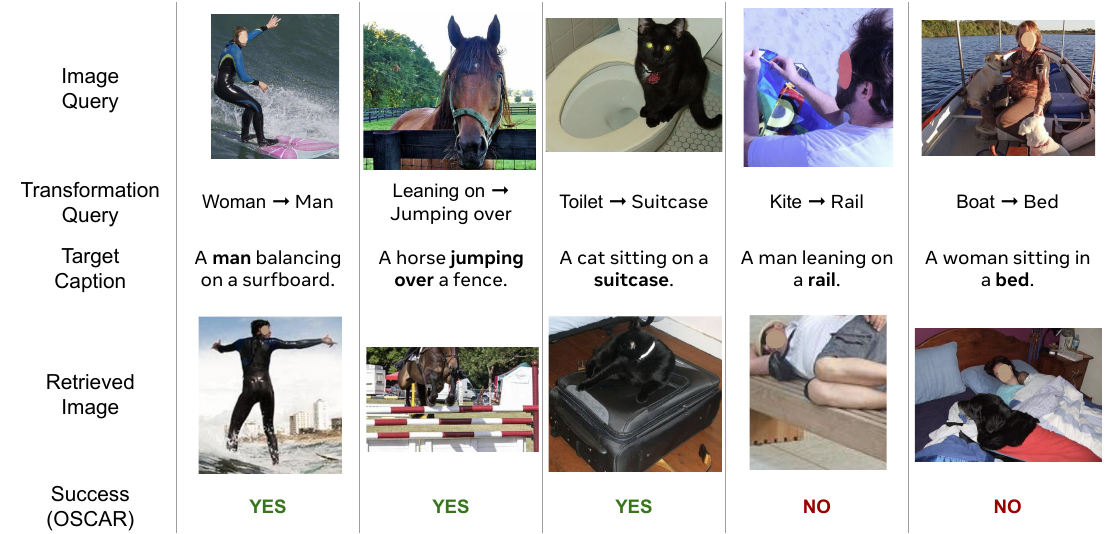}
    \caption{Transformation examples from the CLIP model finetuned on COCO with temperature $\tau=0.1$. Rows 1-3 show examples of successful subject, relation and object transformations. Row 4 shows an example of an unsuccessful object transformation: the retrieved image contains a bench instead of a rail, but we can note some visual similarity with a rail. Row 5 shows a frequent mode of failure: the object is the correct one but the relation has been modified. We assume that our algorithm prioritized keeping the dog in the image.}
    \label{fig:examples}
\end{figure*}

\subsection{Vanilla CLIP embeddings}
\label{sec:zsclip}

We first study the performance of the vanilla CLIP embeddings for transferring text transformations to images with delta vectors. To put our results in perspective, we also evaluate the following baselines:

    \textbf{Text to Image}: We directly provide the target captions to the CLIP text encoder and retrieve the image closest to that embedding. This is the standard image retrieval task, which is easier because the target subject-relation-object features are given as input. Hence it can be considered as an upper bound of our SIMAT score.

    \textbf{Image to Text to Image}: We first find among the SIMAT captions, the text embedding that is closest to the query image. We then add the text delta vector corresponding to the transformation query and finally retrieve the closest image in the SIMAT database. This means that we do not transform the input image directly but we transform a textual representation of the input image.
    

Results are shown in Table~\ref{clip_zs}.
The delta vector method works for 15.9\% of the transformation queries. A higher value of $\lambda$ gives much better results (35.4\%) which are nonetheless below the Image to Text to Image baseline (39\%), and very far from the Text to Image upper bound (65.9\%).
It means that with our benchmark, transforming images works better by using text representations of images rather than the image embeddings themselves. However, in a real-world scenario, we don't want to get explicit context of images by converting them to text (which requires a form of image captioning); we want to use the image embeddings as implicit context.

\begin{table}
\centering
\begin{tabular}{ccc}
\toprule
\textbf{Method} & \multicolumn{2}{c}{SIMAT score} \\
 & $n=1$ & $n=5$ \\
\midrule
\textbf{Delta Vectors ($\lambda=1$)} & 15.9 & 39.2 \\
\textbf{Delta Vectors ($\lambda=3$)} & 35.4 & 67.6 \\
\textbf{Image to Text to Image} & 39.7 & 71.0 \\
\textbf{Text to Image} & 65.9 & 95.6 \\
\bottomrule
\end{tabular}
\caption{\label{clip_zs}
SIMAT score for delta vectors in the original CLIP multimodal space. The default score considers the nearest neighbour in the retrieval step ($n=1$). We also report the SIMAT score for the the best image using $n=5$ nearest neighbours.
}

\end{table}

\subsection{Fine-tuning CLIP on COCO}
\label{sec:ftcoco}

\begin{figure}[H]
    \centering
    \includegraphics[width=1.0\linewidth]{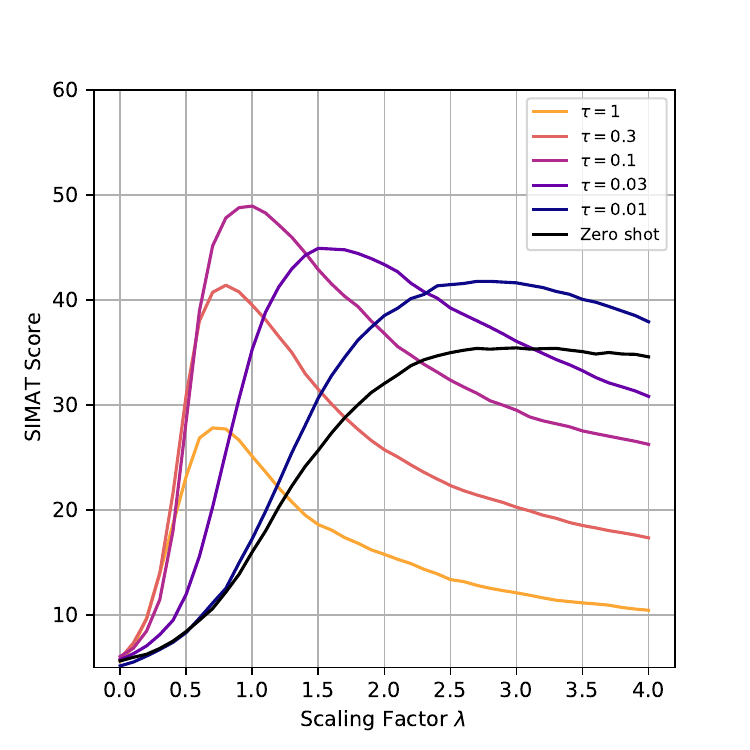}
    \caption{SIMAT score as a function of the scaling factor $\lambda$ (on development set). The overall best score is obtained for $\tau=0.1$
    and a scaling factor of exactly 1.}
    \label{fig:temp}
\end{figure}

In this section, we consider CLIP as image and text encoder, but we additionally train adaptation layers on COCO with different values for the temperature parameter $\tau$.
Figure \ref{fig:temp} shows the SIMAT score as a function of the scaling factor $\lambda$, on the SIMAT dev set. The same curve for the vanilla CLIP embeddings is shown in black. 
We can see that all curves have an optimal value for $\lambda$, which depends on $\tau$. This optimal value $\lambda^*(\tau)$ decreases as $\tau$ increases from 0.01 to 1, and the global optimum is reached for $\tau=0.1$ and $\lambda=1$.
For these values, the SIMAT score is 48.2 which is a 33\% absolute improvement over the zero-shot score.

We therefore conclude that the temperature parameter $\tau$ has a great importance for transferring text delta vectors to images, and that the fine-tuned embeddings work best with delta vectors for $\tau=0.1$ and $\lambda=1$. 

Here, we make the important observation that the empirical optimal value for $\lambda$ is exactly the theoretical value of $1$ that should be used to transform bag of word embeddings. Given that $\lambda=1$ was suboptimal for vanilla CLIP embeddings, we make the hypothesis that multimodal embeddings that are optimal for $\lambda \neq 1$ can be projected to embeddings better suited for delta vectors (hence having better geometric regularities) that maximize transformation accuracy for $\lambda = 1$.

Transformation examples on SIMAT obtained with this model are presented in Figure \ref{fig:examples}.

Note that the best image retrieval and text retrieval evaluations on COCO are obtained for $\tau=0.01$, which hints towards the fact that smaller temperatures are better for image-text retrieval and higher temperatures ($\tau \sim 0.1$) are more compatible with the delta vector framework.
In the rest of the paper, we use a fixed temperature of $\tau=0.1$.

\subsection{Using pretrained text encoders}

\begin{table}[b!]
\centering
\small
\setlength{\tabcolsep}{1.5pt}
\begin{tabular}[t]{c|c|c|c|c|c|c}
\toprule
& & \multicolumn{2}{c|}{\thead{MS Coco}} & \thead{Text} & & \thead{Retrieval} \\[-3pt]
\thead{Image \\ Encoder} & \thead{Sentence \\ Encoder} & \thead{Text \\ R@1} & \thead{Image \\ R@1}& \thead{delta  \\ vectors} & \thead{SIMAT \\ score} & \thead{upper \\ bound}  \\
\midrule

RN50 &   & 25.4 & 22.3 & 88.0 & 44.5 & 76.2 \\
RN152 & CLIP  & 27.6 & 23.5 & 87.2 &	46.0 &	77.7 \\
CLIP &  & 45.2 & 34.8 & 82.4 & 48.2 & 75.4 \\
\midrule

RN50 &   & 17.6 & 15.2 & 95.3 & 44.6 &	65.6 \\
RN152 & FastText & 19.1 & 16.3 & 95.5 &	46.7 &	68.0 \\
CLIP &  & 28.2 & 21.9 & 94.4 & 47.5 & 70.6 \\
 \midrule

RN50 &   & 18.8 & 16.8 & 91.0 &	38.8 &	66.9  \\
RN152 & LaBSE  & 20.4 & 17.9 & 90.7 &	39.9 &	69.0 \\
CLIP &  & 31.4 & 24.9 & 92.9 & 41.9 & 69.9 \\
\midrule
RN50 &  & 17.0 & 15.4 & 92.1 &	37.0 &	67.0 \\
RN152 & LASER & 19.0 & 16.9 & 92.6 & 	36.0 &	67.6 \\
CLIP &  & 29.6 & 22.8 & 92.8 & 37.7 & 67.6  \\

\midrule
\end{tabular}

\caption{\label{bigtable}
Comparison of different image and sentence encoders for the evaluation of delta vectors ($\tau=0.1$).
}
\end{table}

\begin{figure}[t!]
    \centering
    \includegraphics[width=1.0\linewidth]{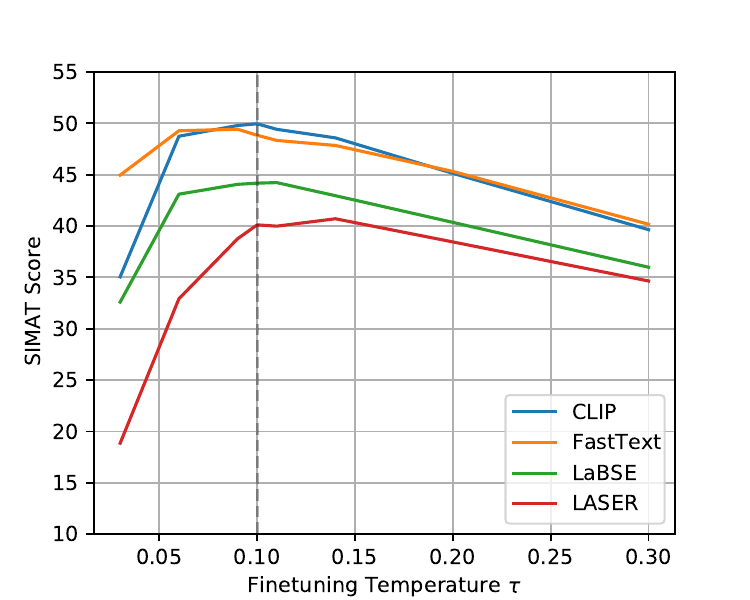}
    \caption{SIMAT score ($\lambda$=1) as a function of training temperature, for several text encoders. For all, the maximum SIMAT score is always obtained for $\tau\approx0.1$.}
    \label{fig:tempfig}
\end{figure}

We show in Figure~\ref{fig:tempfig}, that a value of $\tau=0.1$ which is optimal for CLIP, is also near-optimal for all other considered text embeddings, FastText, LASER and LaBSE. It seems to be a value that works well for delta vectors.
In Table~\ref{bigtable}, we analyse our different choices for the image and text encoders. The \textit{Retrieval upper bound} metric corresponds to the Text to Image baseline of section \ref{sec:zsclip}. The \textit{Text delta vector} metric is an evaluation of how well the text-defined delta vector can accurately transform the caption of the input image (and not the image itself). We also compute the standard image/text retrieval metrics (Image R@1 and Text R@1) on the COCO test set.

We can see that the key contributing factor in the different SIMAT scores is which sentence encoder has been used. If we fix the sentence encoder, the image encoder has an important influence on the image-text retrieval metrics but very little impact on the SIMAT score. Therefore we conclude that improving multimodal embeddings at the task of image/text retrieval will not necessarily improve their geometric properties (in the context of delta vectors).


Also, quite unexpectedly, the SIMAT score does not seem correlated to the Text delta vector score, which measures how well delta vectors can transform text embeddings: the fine-tuned CLIP text embeddings have a text transformation accuracy of 82.4\% whereas the fine-tuned FastText embeddings reach $94.4\%$. Yet they have very similar SIMAT scores (48.2\% vs 47.5\%). It seems to show that within our constraints, a slightly lower performance on text delta vector (which indicates an embedding space with less geometric structure on the text side) is not the current limitation.

\subsection{Sentence-based delta vectors}

In our default method for using text-based delta vectors, we used single words as input to the text encoder. This is particularly well suited for the FastText embeddings which are based on word embeddings, but not so much for the LASER and LaBSE sentence encoders which are built to encode sentences and not single words. This could explain the performance gap between FastText and LASER/LaBSE. To test this hypothesis, we changed our definition of delta vectors so that it is computed by encoding sentences rather than single words.
%
%
We define the \textit{sentence average delta vector} of transformation $\wi$ as the average of delta vectors $E(s_2) - E(s_1)$ where $s_1$ and $s_2$ go over all pairs of SIMAT captions such that $s_2$ is the result of the text transformation $\wi$ applied to $s_1$.

We show the results in Table~\ref{fig:avgtbl}. With this new method, the performance gap between the different text encoders is much smaller, the SIMAT score being higher for LASER and LaBSE, and smaller for FastText. We observed that we can use a higher scaling factor to boost the SIMAT score, up to $\lambda=1.5$ for CLIP. We suspect this is due to the fact that the second method produces more reliable delta vectors with a smaller norm.

Note that the role of this experiment is to shed light on the reasons behind the performance spread with respect to the text encoders. The captions of SIMAT should be reserved for evaluation only and not used within the algorithm.
A better algorithm may use the COCO captions to create better sentence-based delta vectors, but we leave this for future work.

\begin{table}
\centering

\begin{tabular}{c|c|ccc}
\toprule
\thead{Sentence \\ Encoder} & \thead{Single \\ word} & \multicolumn{3}{c}{\thead{Sentence \\ Average}}  \\
 & & $\lambda=1$ & $\lambda=1.2$ & $\lambda=1.5$ \\
\midrule

CLIP & 48.2 & 46.7 & 51.5 & 53.5 \\
FastText & 47.5 &
44.6 & 46.5 & 45.8 \\
LASER & 37.7 & 43.8 & 45.0 & 44.2 \\
LaBSE & 41.9 & 44.6 & 46.5 & 45.5 \\
\bottomrule
\end{tabular}
\caption{Comparison of two methods to calculate delta vectors: \textit{Single word} and \textit{Sentence average}. With the latter, all the encoders have very similar SIMAT scores.}
    \label{fig:avgtbl}
\end{table}

%% file: conclusion.tex
\section{Discussion}

In this paper, we introduce SIMAT, a novel dataset to study the task of text-driven image transformation. It is much larger in size and variety of transformations than existing approaches like SIC112. Due to this larger size, we argue that evaluation cannot be performed solely by using the caption of the retrieved image, and we propose to use OSCAR to assess whether an input image has been successfully transformed.

We use SIMAT to study the geometric properties of multimodal embedding spaces trained with an image-text alignment objective. We use a simple linear approach (\textit{delta vectors}) for transferring text-defined transformations to images in multimodal spaces, which should work well for well-structured spaces. This provides a novel way to study multimodal embedding spaces compared to standard image/text retrieval metrics in the litterature.

After having evaluated vanilla CLIP multimodal embeddings, we have studied embeddings obtained by training with an image/text alignment on COCO, that use pretrained text encoders (FastText, LASER, LaBSE) and pretrained image encoders (CLIP, Resnet50, Resnet152). We emphasize below our findings:

\begin{itemize}
    \item Vanilla CLIP embeddings, although very powerful for image/text retrieval, are not very well suited for delta-vector based transformation (Tab. \ref{clip_zs}).
    \item Finetuning CLIP on COCO brings substantial improvements for delta-vector based transformations and the best performance is obtained for $\tau=0.1$ and $\lambda=1$ (Fig. \ref{fig:temp} We also observe that $(\tau=0.1, \lambda=1)$ is the best functioning point for all considered pretrained text encoders (FastText, LASER, LaBSE, Fig. \ref{fig:tempfig}). Since $\lambda=1$ is also the optimal theoretical value with the delta vector framework, we conclude that finetuning at $\tau=0.1$ helps to improve the geometric properties of the multimodal embedding space.
    \item We did not find any evidence that using geometric properties of pretrained sentence embeddings was helpful. While we expected multimodal embedding spaces built on top of these well-behaved text spaces to display better linear properties, experiments have showed the opposite : (a) higher accuracy for text transformation is not correlated to better image transformation (Tab. \ref{bigtable}); (b) Using LASER and LabSE was actually harmful (Fig. \ref{fig:tempfig}) but we show in Tab. \ref{fig:avgtbl} that this is almost entirely due to the fact that we only have access to single words to compute the text delta vector.
    
\end{itemize}

For future work, we would like to extend SIMAT to richer semantic transformations: 
we expect that for transformations like \transfo{young}{old} or \transfo{dirty}{clean}, higher-level semantic knowledge embedded in language models will be critical to do meaningful transformations.